%% file: icra2017_mod.tex
\pdfoutput=1

\documentclass[letterpaper, 10 pt, conference]{ieeeconf}  % Comment this line out if you need a4paper
\IEEEoverridecommandlockouts % makes \thanks work :)
\usepackage[utf8]{inputenc}
\usepackage[T1]{fontenc}
\usepackage{standalone}
\usepackage{pgfplots}
\usepackage{graphicx}
\usepackage{subfigure}
\usepackage{amssymb}
\usepackage{amsmath}
\pgfplotsset{compat=1.14}

\usepackage[capitalise]{cleveref} % Clever references
\crefname{section}{Sec.}{Sections}
\Crefname{section}{Section}{Sections}

\usepackage{url}

\usepackage[keeplastbox]{flushend} %flushend messes with last reference line. keeplastbox fixes it

\usepackage{rotating}

\usepackage{url}

\usepackage[detect-all]{siunitx}
\usepackage{tikz}
\usepgfplotslibrary{groupplots}
\usetikzlibrary{plotmarks}
\usetikzlibrary{positioning}
\usetikzlibrary{shapes,arrows, fit, positioning}
\usepgfplotslibrary{external}
\usetikzlibrary{shapes, fit, arrows}
\usepackage{tikz-3dplot}
\usepackage{tikzscale}
\usepackage{centeredtikzarrowbulletheads}
\tikzset{>=latex}
\tikzset{
  font={\fontsize{10pt}{12}\selectfont}}

\DeclareMathOperator{\dis}{d}

\title{\LARGE \bf
Autonomous Interpretation of Demonstrations for Modification of Dynamical Movement Primitives
}

\author{
\centering Martin Karlsson* \quad Anders Robertsson \quad Rolf Johansson
\thanks{* The authors work at the Department of Automatic Control, \protect\\
Lund University, PO Box 118,
SE-221 00 Lund, Sweden.\protect\\
{Martin.Karlsson@control.lth.se\protect\\
The authors would like to thank Fredrik Bagge Carlson, Bj{\"o}rn~Olofsson and Karl Johan {\AA}str{\"o}m at the Department of Automatic Control, Lund University, as well as Maj~Stenmark, Mathias~Haage and Jacek~Malec at Computer Science, Lund University, for valuable discussions throughout this work. The authors are members of the LCCC Linnaeus Center and the ELLIIT Excellence Center at Lund University. The research leading to these results has received funding from the European Community’s Framework Programme Horizon 2020 – under grant agreement No 644938 – SARAFun.}}
}

\begin{document}
% Some global setup for tikz
\newlength\figureheight 
\newlength\figurewidth 
\newcommand{\cmt}[1]{{\color{red}{\textbf{Comment:} #1}}}
\pgfplotsset{every tick label/.append style={font=\footnotesize}}
\pgfplotsset{legend style={font=\footnotesize}}
\pgfplotsset{every axis label/.append style={font=\footnotesize}}

\maketitle
\thispagestyle{empty}
\pagestyle{empty}

%%%%%%%%%%%%%%%%%%%%%%%%%%%%%%%%%%%%%%%%%%%%%%%%%%%%%%%%%%%%%%%%%%%%%%%%%%%%%%%%
\begin{abstract}
The concept of dynamical movement primitives (DMPs) has become popular for modeling of motion, commonly applied to robots. This paper presents a framework that allows a robot operator to adjust DMPs in an intuitive way. Given a generated trajectory with a faulty last part, the operator can use lead-through programming to demonstrate a corrective trajectory. A modified DMP is formed, based on the first part of the faulty trajectory and the last part of the corrective one. A real-time application is presented and verified experimentally.
\end{abstract}

%%%%%%%%%%%%%%%%%%%%%%%%%%%%%%%%%%%%%%%%%%%%%%%%%%%%%%%%%%%%%%%%%%%%%%%%%%%%%%%%
\section{Introduction}
High cost for time-consuming robot programming, performed by engineers, has become a key obstruction in industrial manufacturing. This has promoted the research towards faster and more intuitive means of robot programming, such as learning from demonstration, of which an introduction is presented in~\cite{argall2009survey}. It is in this context desirable to make robot teaching available to a broader group of practitioners by minimizing the engineering work required during teaching of tasks.

A costumary way to quickly mediate tasks to robots is to use lead-through programming, while saving trajectory data so that the robot can reproduce the motion. In this paper, the data are used to form dynamical movement primitives (DMPs). Early versions of these were presented in \cite{ijspeert2002humanoid}, \cite{schaal2003computational} and \cite{ijspeert2003learning}, and put into context in \cite{niekum2015learning}. Uncomplicated modification for varying tasks was emphasized in this literature. For example, the time scale was governed by one parameter, which could be adjusted to fit the purpose. Further, the desired final state could be adjusted, to represent a motion similar to the original one but to a different goal. DMPs applied on object handover with moving targets were addressed in \cite{prada2014handover}. The scalability in space was demonstrated in, \emph{e.g.}, \cite{ijspeert2013dynamical}.

\begin{figure}
	%\centering	
	%\footnotesize
	%\input{figs/adjust_cart.tex}
	\includegraphics[width=\columnwidth]{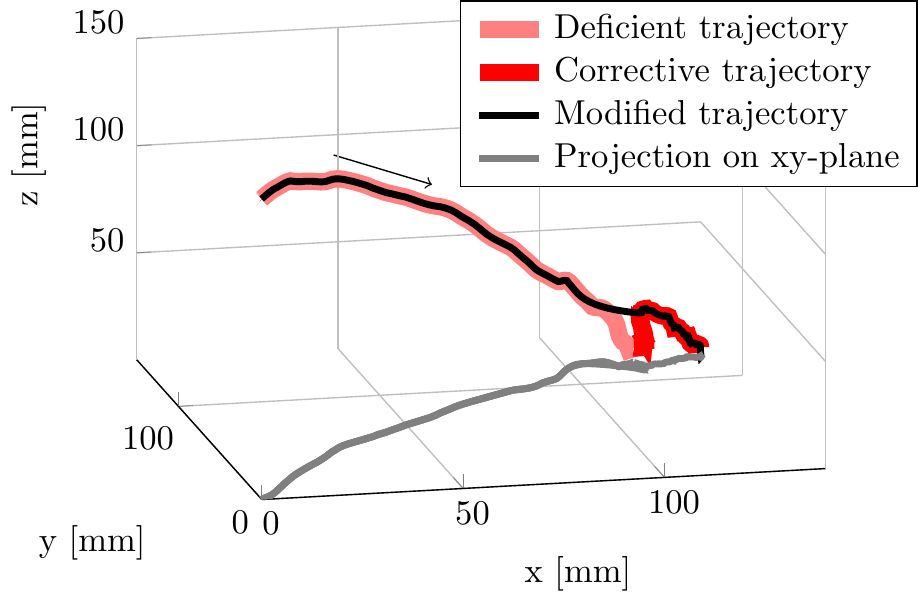}
	\caption{Trajectories of the robot's end-effector from one of the experiments. The arrow indicates the direction. The deficient trajectory was generated from the original DMP. After that, the operator demonstrated the corrective trajectory. Merging of these, resulted in the modified trajectory. The projection on the xy-plane is only to facilitate the visualization.} \label{fig:adjust_cart}
\end{figure}

The scenario considered in this paper is the unfavorable event that the last part of the motion generated by a certain DMP is unsatisfactory. There might be several reasons for this to occur. In the case where the starting points differ, the generated trajectory would still converge to the demonstrated end point, but take a modified path, where the modification would be larger for larger differences between the starting points. Further, the DMP might have been created in a slightly different setup, \emph{e.g.}, for a different robot or robot cell. There might also have been a mistake in the teaching that the operator would have to undo. If the whole last part of the trajectory is of interest, it is not enough to modify the goal state only. One way to solve the problem would be to record an entirely new trajectory, and then construct a corresponding DMP. However, this would be unnecessarily time consuming for the operator, as only the last part of the trajectory has to be modified. Instead, the method described here allows the operator to lead the manipulator backwards, approximately along the part of the trajectory that should be adjusted, followed by a desired trajectory, as visualized in \cref{fig:adjust_cart}. 

\begin{figure}
\includegraphics[width=\columnwidth]{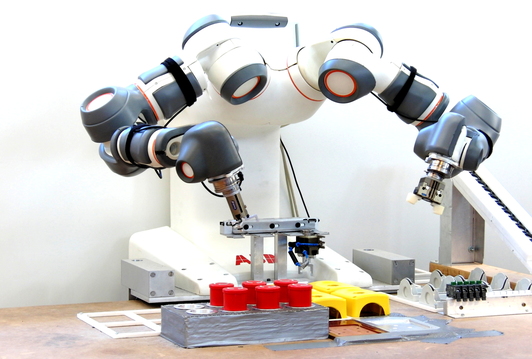}
\caption{The ABB YuMi prototype robot used in the experiments. }\label{fig:yumi5}
\end{figure}

Hitherto, DMPs have usually been formed by demonstrations to get close to the desired behavior, followed by trajectory-based reinforcement learning, as presented in, \emph{e.g.}, \cite{pastor2013dynamic,kober2008learning,abu2015adaptation,kroemer2010combining}. Compared to such refinements, the modification presented here is less time consuming and does not require engineering work. On the other hand, the previous work on reinforcement learning offers modulation based on sensor data, and finer  movement adjustment. Therefore, the framework presented in this paper forms an intermediate step, where, if necessary, a DMP is modified to prepare for reinforcement learning, see \cref{fig:workflow}. This modification can be used within a wide range of tasks. In this paper, we exemplify by focusing on peg-in-hole tasks.

%\end{document}

In \cite{pastor2013dynamic}, online modulation, such as obstacle avoidance, was implemented for DMPs. This approach has been verified for several realistic scenarios, but requires an infrastructure for obstacle detection, as well as some coupling term parameters to be defined. It preserves convergence to the goal point, but since the path to get there is modified by the obstacle avoidance, it is not guaranteed to follow any specific trajectory to the goal. This is significant for, \emph{e.g.}, a peg-in-hole task.

The paper is outlined as follows. Two example scenarios in which the framework would be useful are presented in \cref{sec:motivexamples}, followed by a description of the method in \cref{sec:framework}. Experimental setup and results are described in \cref{sec:experiments,sec:results}, and finally a discussion and concluding remarks are presented in \cref{sec:discussion,sec:conclusions}, respectively.

\section{Problem Formulation}
\label{sec:problemformulation}
In this paper, we address the question whether it is possible to automatically interpret a correction, made by an operator, of the last part of a DMP trajectory, while still taking advantage of the first part. The human-robot interaction must be intuitive, and the result of a correction predictable enough for its purpose. The correction should result in a new DMP, of which the first part behaves qualitatively as the first part of the original DMP, whereas the last part resembles the last part of the corrective trajectory. Any discontinuity between the original and corrective trajectories must be mitigated.

%\section{Notation}
%For convenience, we here provide a list of some of the more important quantities used in this paper:

%\begin{table}[!hbt]
%\normalsize
%\begin{center}
%\begin{tabular}{l l}
%$\tau$ & - Time constant of a DMP\\
%$g$ & - Goal (attractor) of a DMP\\
%$f(x)$ & - Nonlinear term of a DMP\\
%$\Psi_i$ & - Basis function $i$ of nonlinear term\\
%$\omega_i$ & - Weight of basis function $i$\\
%$y_{d}$ & - Deficient trajectory\\
%$y_{c}$ & - Corrective trajectory\\
%$y_{cr}$ & - Retained part of corrective trajectory\\
%$y_{dr}$ & - Retained part of deficient trajectory\\
%$y_{m}$ & - Modified part of $y_{dr}$\\
%$\dis(a,b)$ & - Euclidean distance between point $a$ and $b$ \\
%$notation_1$ & - explanation \\
%$notation_2$ & - explanation \\
%\end{tabular}
%\end{center}
%\end{table}
%This notation will be explained in more detail later on, while this list may serve as a short reference.

\section{Motivating Examples}
\label{sec:motivexamples}
We here describe two scenarios where the framework proves useful. These are evaluated in \cref{sec:experiments,sec:results}, where more details are given.

\subsection{Inadequate precision - Scenario A}
\label{sec:inadequate}
\begin{figure}
\subfigure[\label{fig:pm1}]{\includegraphics[height=3cm]{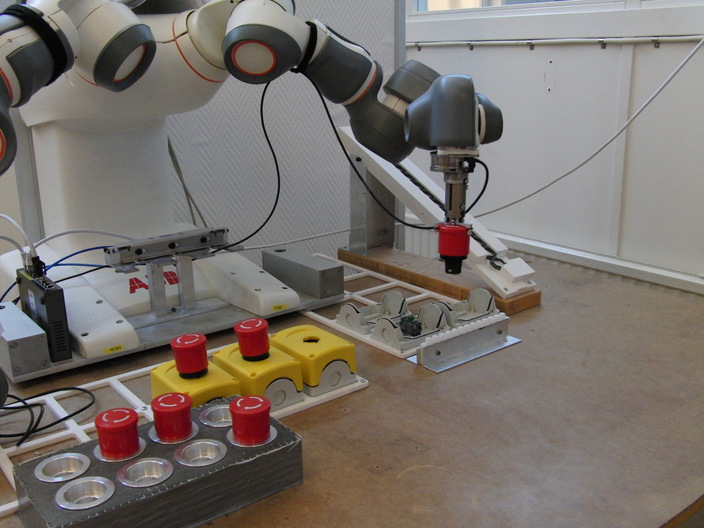}}\hfill%
%\vspace{-.6\baselineskip}
\subfigure[\label{fig:pm2}]{\includegraphics[height=3cm]{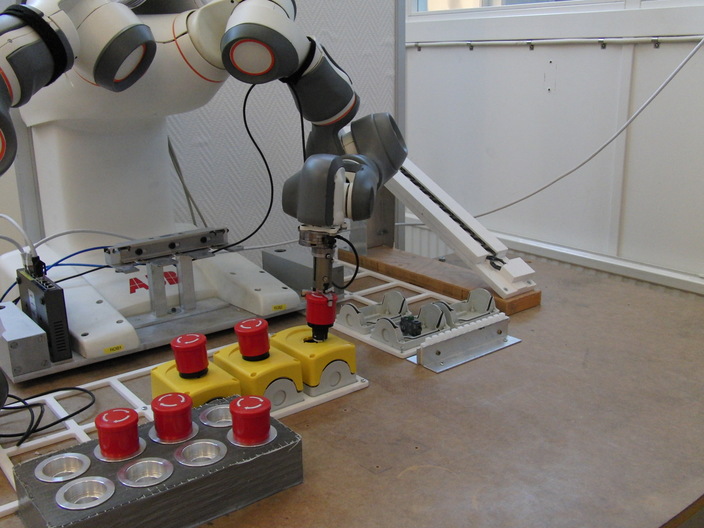}}
\vspace{-.6\baselineskip}
\subfigure[\label{fig:pm3}]{\includegraphics[height=3cm]{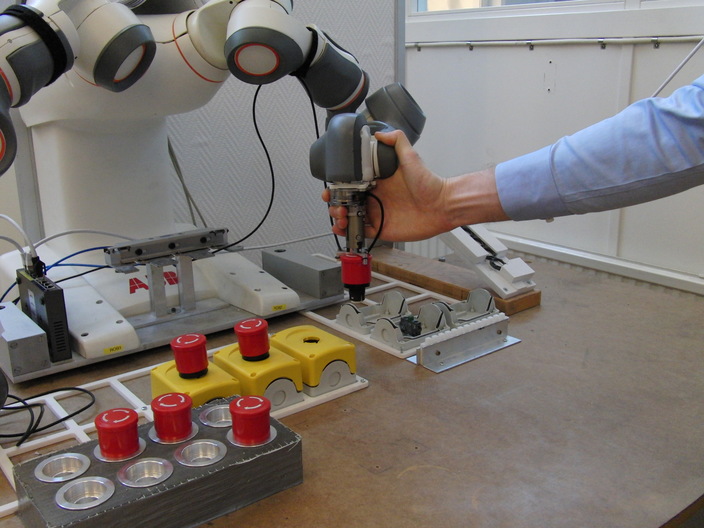}}
\hfill%
%\vspace{-.6\baselineskip}
\subfigure[\label{fig:pm4}]{\includegraphics[height=3cm]{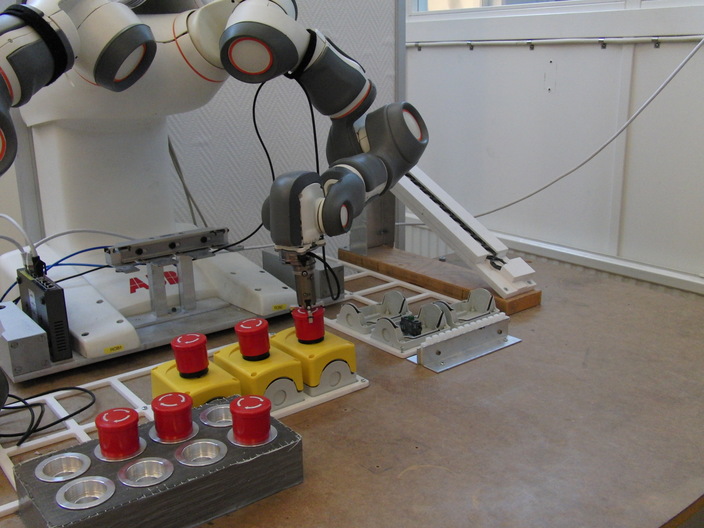}}
\caption{Scenario A. The evaluation started in (a), and in (b) the robot failed to place the button in the hole due to inadequate accuracy. Between (a) and (b), the deficient trajectory was recorded. The operator led the robot arm (c) backwards, approximately along a proportion of the deficient trajectory, and subsequently led it to place the button properly, while the corrective trajectory was recorded. The robot then made the entire motion, starting in a configuration similar to that in (a), and ending as displayed in (d).}
\label{fig:pm}
\end{figure}

Consider the setup shown in \cref{fig:pm}, where the button should be placed into the yellow case. A DMP was run for this purpose, but, due to any of the reasons described above, the movement was not precise enough, and the robot got stuck on its way to the target, see \cref{fig:pm2}. Hitherto, such a severe shortcoming would have motivated the operator to teach a completely new DMP, and erase the old one. With the method proposed here, the operator had the opportunity to approve the first part of the trajectory, and only had to modify the last part. This was done by leading the robot arm backwards, approximately along the faulty path, until it reached the acceptable part. Then, the operator continued to lead the arm along a desired path to the goal. When this was done, the acceptable part of the first trajectory was merged with the last part of the corrective trajectory. After that, a DMP was fitted to the resulting trajectory. Compared to just updating the target point, this approach also allowed the operator to determine the trajectory leading there. This scenario is referred to as Scenario~A.

\subsection{New obstacle - Scenario B}
\label{sec:novelobstacle}
\begin{figure}
	\subfigure[\label{fig:as1}]{\includegraphics[height=3cm]{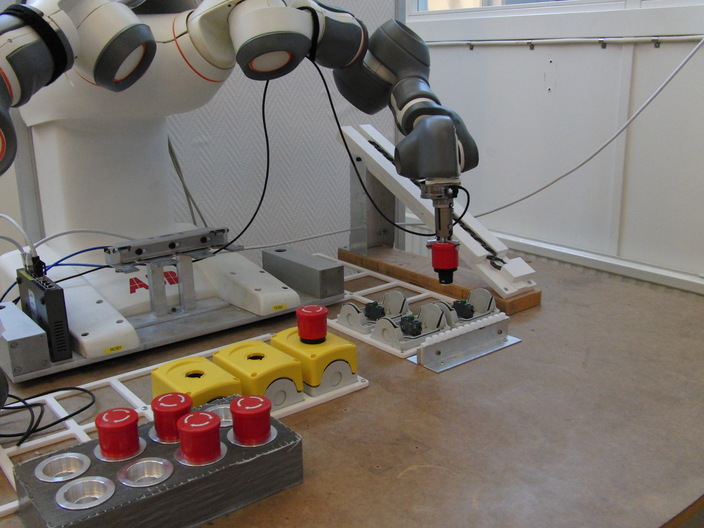}}\hfill%
	%\vspace{-.6\baselineskip}
	\subfigure[\label{fig:as2}]{\includegraphics[height=3cm]{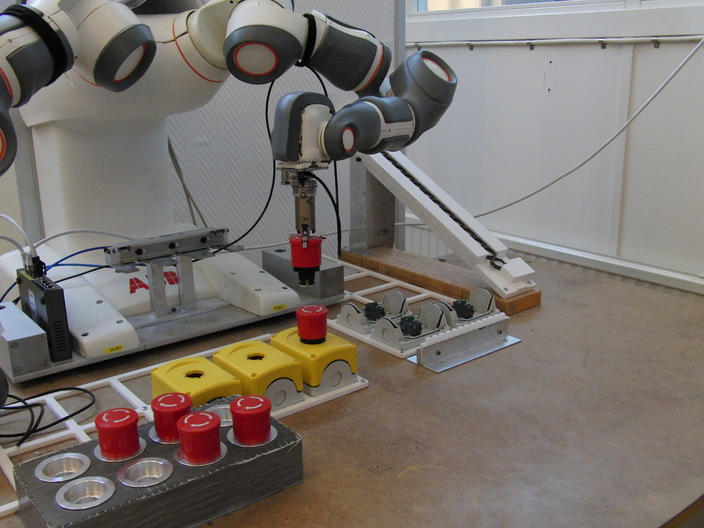}}
	\subfigure[\label{fig:as3}]{\includegraphics[height=3cm]{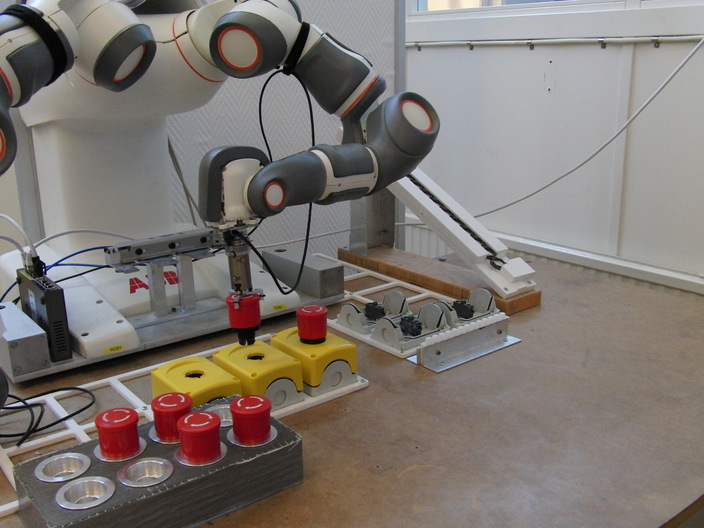}}
	\hfill%
	\subfigure[\label{fig:as4}]{\includegraphics[height=3cm]{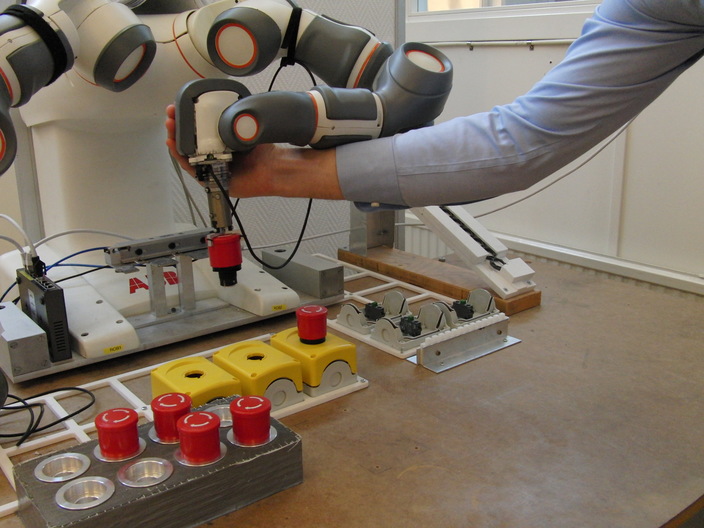}}
	\subfigure[\label{fig:as5}]{\includegraphics[height=3cm]{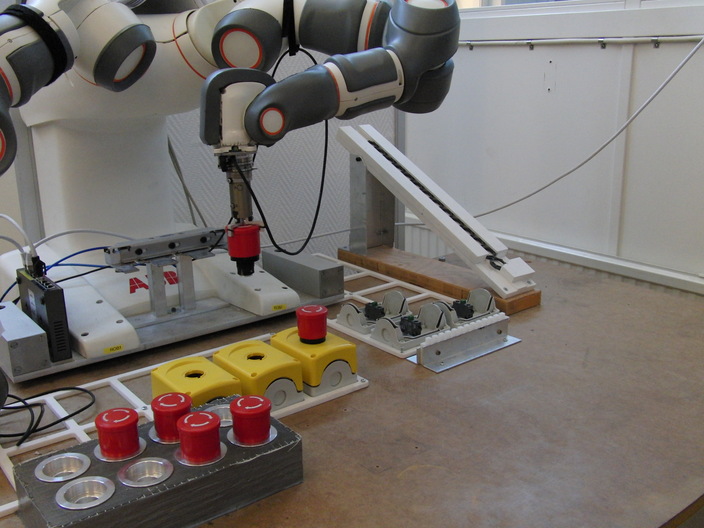}}
	\hfill%
	%\vspace{-.6\baselineskip}
	\subfigure[\label{fig:as6}]{\includegraphics[height=3cm]{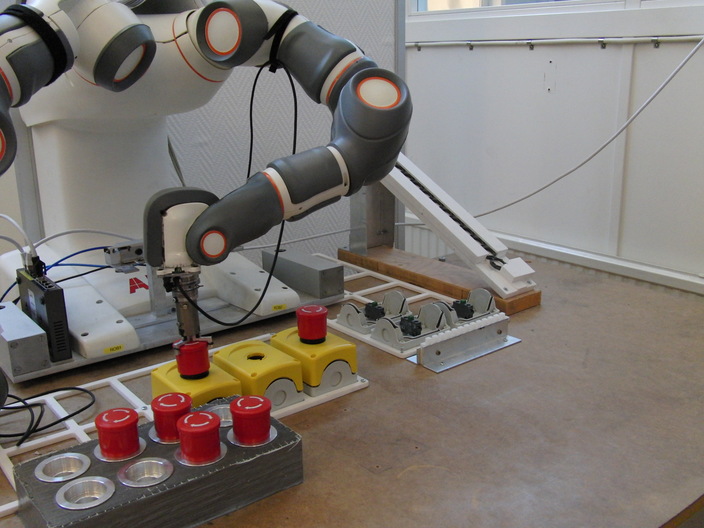}}
	\caption{Scenario B. The initial goal was to move the button to the leftmost yellow case, above the hole, to prepare for placement. The evaluation started in (a), and in (b) the trajectory was satisfactory as the placed button was avoided. In (c), however, there would have been a collision if there was a button placed in the middle case. Further, it was desired to complete the peg-in-hole task, rather than stopping above the hole. Hence, the evaluated trajectory was considered deficient. In (d), the operator led the robot arm back, and then in a motion above the potential obstacle, and into the hole, forming the corrective trajectory. Based on the modified DMP, the robot started in a position similar to that in (a), avoided the potential obstacle in (e) and reached the new target in (f). The trajectories from one attempt are shown in \cref{fig:avoid1}. }
	\label{fig:as}
\end{figure}

For the setup in \cref{fig:as}, there existed a DMP for moving the robot arm from the right, above the button that was already inserted, to a position just above the hole in the leftmost yellow case. However, under the evaluation the operator realized that there would have been a collision if a button were already placed in the case in the middle. A likely reason for this to happen would be that the DMP was created in a slightly different scene, where the potential obstacle was not taken into account. Further, the operator desired to extend the movement to complete the peg-in-hole task, rather than stopping above the hole. With the method described herein, the action of the operator would be similar to that described in~\cref{sec:inadequate}, again saving work compared to previous methods. This scenario is referred to as Scenario~B.

\section{Description of the Framework}
\label{sec:framework}

\begin{figure}
	\centering
    \input{./figs/cascade_scheme.tex}
    \caption{Schematic visualization of the work flow, from an operator's perspective. A DMP was created based on a demonstration. Subsequently, the DMP was executed while evaluated by the operator. If unsuccessful, the operator demonstrated a correction, which yielded a modified DMP to be evaluated. Once successful, further improvement could be done by, \emph{e.g.}, trajectory-based reinforcement learning, though that was outside the scope of this work. Steps that required direct, continuous interaction by the operator are marked with light {\color{red} red} color. Steps that required some attention, such as supervision and initialization, are marked with light {\color{blue} blue}. The operations in the white boxes were done by the software in negligible computation time, and required no human involvement. The work in this paper focused on the steps within the dashed rectangle.}
\label{fig:workflow}
\end{figure}
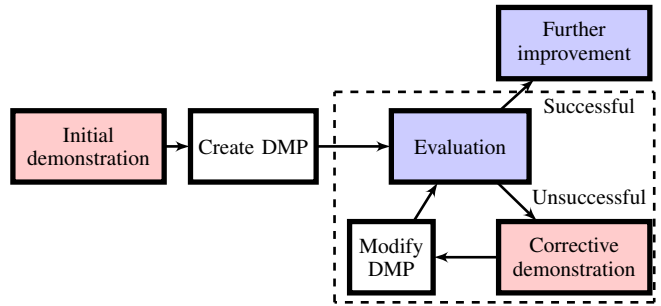

In this section, the concept of DMPs is first introduced. A method to determine what parts of the deficient and corrective trajectories to retain is presented, followed by a description of how these should be merged to avoid discontinuities. Finally, some implementation aspects are addressed. \Cref{fig:workflow} displays a schematic overview of the work flow of the application, from the user's perspective.

\subsection{Dynamical movement primitives}
A review of the DMP concept was presented in~\cite{ijspeert2013dynamical}, and here follows a short description of how it was applied in this context. A certain trajectory, $y$, was modeled by the system
\begin{align}
\tau \dot{y} = z,
\label{eq:getting_ydot}
\end{align}
where $z$ is determined by
\begin{align}
\tau \dot{z} =& \alpha_z(\beta_z(g-y)-z) + f(x)
\label{eq:getting_zdot}
\end{align}
In turn, $f(x)$ is a function given by
\begin{align}
f(x) =& \frac{\sum_{i=1}^{N_b} \Psi_i(x)w_i}{\sum_{i=1}^{N_b} \Psi_i(x)} x \cdot (g-y_0),
\end{align}
where the basis functions, $\Psi_i(x)$, take the form
\begin{align}
\Psi_i(x) =& \exp \left(-\frac{1}{2\sigma_i^2}(x-c_i)^2 \right) \\
\tau \dot{x} =& -\alpha_x x
\end{align}
Here, $\tau$ is a time constant, while $\alpha_z$, $\beta_z$ and $\alpha_x$ are positive parameters. Further, $N_b$ is the number of basis functions, $w_i$ is the weight for basis function $i$, $y_0$ is the starting point of the trajectory $y$, and $g$ is the goal state. $\sigma_i$ and $c_i$ are the standard deviation and mean of each basis function, respectively. Given a DMP, a robot trajectory can be generated from \cref{eq:getting_ydot,eq:getting_zdot}. Vice versa, given a demonstrated trajectory, $y_{demo}$, a corresponding DMP can be formed; $g$ is then given by the end position of $y_{demo}$, whereas $\tau$ can be set to get a desired time scale. Further, the solution of a weighted linear regression problem in the sampled domain yields the weights

\begin{align}
&w_i = \frac{\boldsymbol{s}^T \boldsymbol{\Gamma}_i \boldsymbol{f}_{target}}{\boldsymbol{s}^T \boldsymbol{\Gamma}_i \boldsymbol{s}},
\text{where} \hspace{2mm}
\boldsymbol{s} = 
\begin{pmatrix}
x^1 (g - y_{demo}^1) \\
x^2 (g - y_{demo}^1) \\
\vdots \\
x^{N} (g - y_{demo}^1)
\end{pmatrix}, \\
&\boldsymbol{\Gamma_i} = \text{diag}(\Psi_i^1, \Psi_i^2 \cdots \Psi_i^N), \hspace{2mm}
\boldsymbol{f}_{target} = 
\begin{pmatrix}
f_{target}^1 \\
f_{target}^2 \\
\vdots \\
f_{target}^N
\end{pmatrix}, \\
&f_{target} = \tau^2 \ddot{y}_{demo} - a_z(b_z(g-y_{demo}) - \tau \dot{y}_{demo}).
\end{align}
Here, $N$ is the number of samples in the demonstrated trajectory.

\subsection{Interpretation of corrective demonstration}
\label{sec:interpret}
\begin{figure}
	\centering	
	\setlength{\figurewidth}{0.75\linewidth}
	\setlength{\figureheight}{5cm}
	\footnotesize
	\input{figs/adjust_zoom.tex}
	\caption{Same trajectories as in \cref{fig:adjust_cart}, but zoomed in on the corrective trajectory. Arrows indicate directions. The parts of the trajectories between the separation markers were not retained. The right, {\color{blue} blue}, separation point was determined explicitly by the operator during the corrective demonstration. The left, {\color{green} green}, separation point was determined according to \cref{eq:mindist}. Further, what was left of the deficient trajectory was modified for a smooth transition. However, the part of the corrective trajectory retained was not modified, since it was desired to closely follow this part of the demonstration. Note that the trajectories retained were not intended for direct play-back execution. Instead, they were used to form a modified DMP, which in turn generated a resulting trajectory, as shown in Figs.~\ref{fig:place1}, \ref{fig:place2} and \ref{fig:avoid1}.}  
\label{fig:adjust_zoom}
\end{figure}
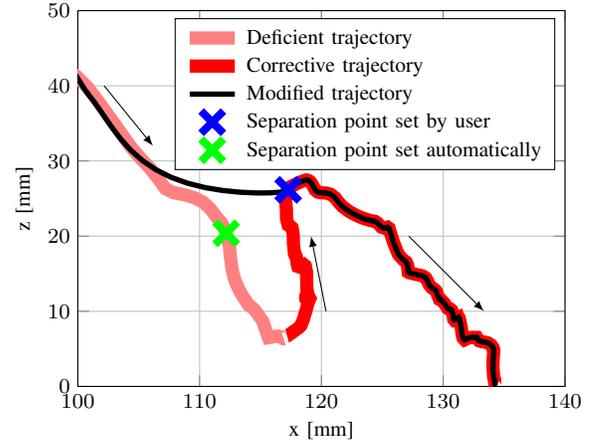

If the evaluation of a trajectory was unsuccessful, a corrective demonstration and DMP modification should follow (\cref{fig:workflow}). Denote by $y_{d}$ the deficient trajectory, and by $y_{c}$ the corrective one, of which examples are shown in Figs.~\ref{fig:adjust_cart}, \ref{fig:adjust_notes} and \ref{fig:adjust_zoom}. A trajectory formed by simply appending $y_{c}$ to $y_{d}$ was likely to take an unnecessary detour. Thus, only the first part of $y_{d}$ and the last part of $y_{c}$ were retained. This is illustrated in \cref{fig:adjust_zoom}. Denote by $y_{cr}$ the retained part of the corrective trajectory. The operator signaled where to separate the corrective trajectory, during the corrective demonstration. In the current implementation, this was done by pressing a button in a terminal user interface, when the robot configuration corresponded to the desired starting point of $y_{cr}$, denoted $y_{cr}^{1}$.

\begin{figure}
	\centering	
	\setlength{\figurewidth}{0.75\linewidth}
	\setlength{\figureheight}{5cm}
	\footnotesize
	\input{figs/adjust_notes.tex}
	\caption{Visualization of shortest distance, here denoted $d_m$, used to determine the left separation marker in \cref{fig:adjust_zoom}. The trajectories are the same as in Figs. \ref{fig:adjust_cart} and \ref{fig:adjust_zoom}, except that the modified trajectory is omitted. }
\label{fig:adjust_notes}
\end{figure}
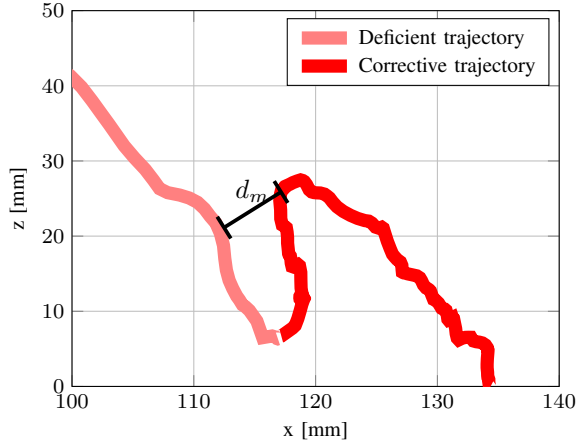

The next step was to determine which part of $y_{d}$ to retain. This was chosen as the part previous to the sample on $y_{d}$ that was closest to $y_{cr}^{1}$, \emph{i.e.},
\newcommand{\argmin}{\operatornamewithlimits{argmin}}
\begin{align}
&y_{dr}^{m} = y_{d}^{m}, \hspace{4mm} \forall m \in [1;M]
\end{align}
where
\begin{align}
&M = \argmin\limits_{k = 1\dots K} \dis(y_{d}^{k}, y_{cr}^1)
\label{eq:mindist}
\end{align}
Here, $\dis$ denotes distance, and $K$ is the number of samples in $y_d$, see \cref{fig:adjust_notes} for an illustration. The approach of using the shortest distance as a criteria, was motivated by the assumption that the operator led the robot arm back, approximately along the deficient trajectory, until the part that was satisfactory. At this point, the operator separated the corrective demonstration, thus defining $y_{cr}^{1}$ (see right marker in \cref{fig:adjust_zoom}). By removing parts of the demonstrated trajectories, a significant discontinuity between the remaining parts was introduced. In order to counteract this, $y_{dr}$ was modified into $y_{m}$, where the following features were desired.
\begin{itemize}
\item $y_{m}$ should follow $y_{dr}$ approximately
\item The curvature of $y_{m}$ should be moderate
\item $y_{m}$ should end where $y_{cr}$ began, with the same direction in this point
\end{itemize}
To find a suitable trade-off between these objectives, the following convex optimization problem was solved:
\begin{align}
& \underset{y_{m}}{\text{minimize}}
& & \lVert y_{dr} - y_{m} \rVert_2 + \lambda 
\lVert T_{(\Delta^2)} y_{m} \rVert_2 \label{eq:opt1}\\
& \text{subject to}
& & y_{m}^{M} = y_{cr}^1 \label{eq:opt2}\\
& & & y_{m}^{M} - y_{m}^{M-1} = y_{cr}^2 - y_{cr}^1 \label{eq:opt3}
\end{align}
Here, $\lambda$ denotes a constant scalar, and $T_{(\Delta^2)}$ is a second-order finite difference operator. Subsequently, $y_{cr}$ was appended on $y_{m}$, and one corresponding DMP was created, with the method described in the previous subsection. The next step in the work flow was to evaluate the resulting DMP (\cref{fig:workflow}). 
\subsection{Software implementation}
\label{sec:implementation}
The research interface ExtCtrl \cite{blomdell2005extending}, \cite{blomdell2010flexible}, was used to send references to the low-level robot joint controller in the ABB IRC5 system \cite{irc5}, at 250 Hz. Most of the programming was done in C++, where DMPs were stored as objects. Among the data members of this class were the parameters $\tau$, $g$ and $w_{1 \dots N_b}$, as well as some description of the context of the DMP and when it was created. It contained member functions for displaying the parameters, and for modifying $g$ and $\tau$. The communication between the C++ program and ExtCtrl was handled by the LabComm protocol \cite{labcomm}. The C++ linear algebra library Armadillo \cite{sanderson2010armadillo} was used in a major part of the implementation. Further, the code generator CVXGEN \cite{mattingley2012cvxgen} was used to generate C code for solving the optimization problem in Eqs. (\ref{eq:opt1}), (\ref{eq:opt2}) and (\ref{eq:opt3}). By default, the solver code was optimized with respect to computation time. This resulted in a real-time application, in which the computation times were negligible in teaching scenarios. The optimization problem was typically solved well below one millisecond on an ordinary PC.

\section{Experiments}
\label{sec:experiments}
The robot used in the experimental setup was a prototype of the dual-arm ABB YuMi \cite{yumi} (previously under the name FRIDA) robot, with 7 joints per arm, see \cref{fig:yumi5}. The experiments were performed in real-time using the implementation described in \cref{sec:implementation}. The computations took place in joint space, and the robot's forward kinematics was used for visualization in Cartesian space in the figures presented. The scenarios in \cref{sec:motivexamples} were used to evaluate the proposed method. For each trial, the following steps were taken. 
\begin{itemize}
\item An initial trajectory was taught, deliberately failing to meet the requirements, as explained in \cref{sec:motivexamples}.
\item Based on this, a DMP was created.
\item The DMP was used to generate a trajectory similar to the initial one. This formed the deficient trajectory.
\item A corrective trajectory was recorded.
\item Based on the correction, a resulting DMP was formed automatically.
\item The resulting DMP was executed for experimental evaluation.
\end{itemize}

First, Scenario A was set up for evaluation, see \cref{sec:inadequate} and \cref{fig:pm}. The scenario started with execution of a deficient trajectory. For each attempt, a new deficient trajectory was created and modified. A total of 50 attempts were made. 

Similarly, Scenario B (see \cref{sec:novelobstacle} and \cref{fig:as}) was set up, and again, a total of 50 attempts were made.

A video is available as a publication attachment, to facilitate understanding of the experimental setup and results. A version with higher resolution is available on \cite{mod_youtube}.

\section{Results}
\label{sec:results}
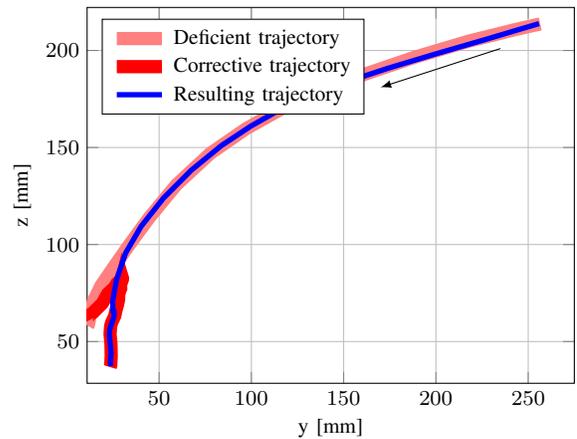
\begin{figure}
	\centering	
	\setlength{\figurewidth}{0.75\linewidth}
	\setlength{\figureheight}{5cm}
	\footnotesize
	\input{figs/place_a1.tex}
	\caption{Trajectories from the experimental evaluation of Scenario A. The deficient trajectory went past the goal in the negative $y$-direction,  preventing the robot from lowering the button into the hole. After correction, the robot was able to reach the target as the modified DMP generated the resulting trajectory.} \label{fig:place1}
\end{figure}

\begin{figure}
	\centering	
	\setlength{\figurewidth}{0.75\linewidth}
	\setlength{\figureheight}{5cm}
	\footnotesize
	\input{figs/place_a2.tex}
	\caption{Similar to \cref{fig:place1}, except that in this case, the deficient trajectory did not reach far enough in the negative $y$-direction.} \label{fig:place2}
\end{figure}
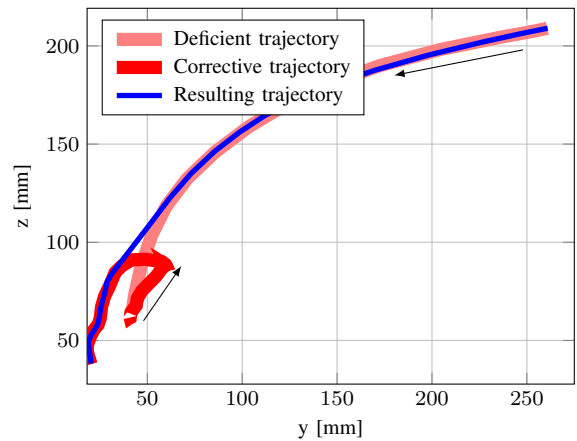

For each attempt of Scenario A, the robot was able to place the button properly in the yellow case after modification. Results from two of these attempts are shown in Figs.~\ref{fig:place1} and \ref{fig:place2}. In the first case, the deficient trajectory went past the goal, whereas in the second case, it did not reach far enough.

% PLACE PLOTS

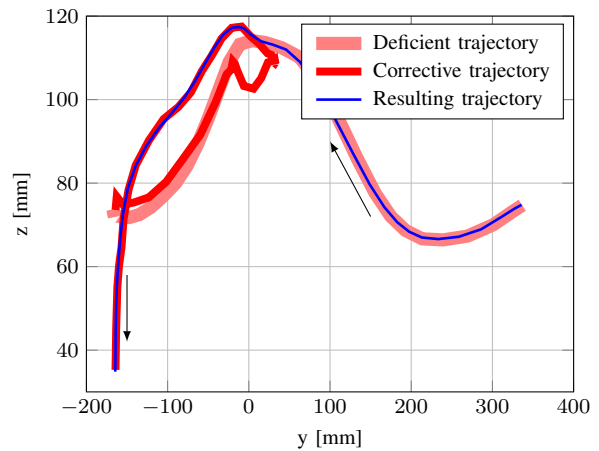
\begin{figure}
	\centering	
	\setlength{\figurewidth}{0.75\linewidth}
	\setlength{\figureheight}{5cm}
	\footnotesize
	\input{figs/avoid_a1.tex}
	\caption{Trajectories from experimental evaluation of Scenario B. The deficient trajectory was lowered too early, causing a potential collision. After the correction, the robot was able to reach the target while avoiding the obstacles. The movement was also extended to perform the entire peg-in-hole task, rather than stopping above the hole.} \label{fig:avoid1}
\end{figure}

Each of the attempts of Scenario B were also successful. After modification, the DMPs generated trajectories that moved the grasped stop button above the height of potential obstacles, in this case other stop buttons, and subsequently inserted it into the case. The result from one attempt is shown in  \cref{fig:avoid1}.

\section{Discussion}
\label{sec:discussion}

The subsequent step in this work is to integrate the presented framework with trajectory-based reinforcement learning \cite{pastor2013dynamic}, \cite{stulp2012model}, in order to optimize the motion locally with respect to criteria such as execution time. The program should also be augmented to take the purpose of, and relation between, different DMPs into consideration. This extension will emphasize the necessity of keeping track of different states within the work flow. To this purpose, a state machine implemented in, \emph{e.g.}, JGrafchart \cite{theorin2014sequential}, or the framework of behavior trees, applied on robot control in \cite{marzinotto2014towards}, would be suitable. Extending the user interface with support for natural language, would possibly make this framework more user friendly.

Performing the computations in joint space instead of Cartesian space allowed the operator to determine the entire configuration of the 7~DOF robot arm, rather than the pose of the tool only. However, one could think of situations where the operator is not concerned by the configuration, and the pose of the tool would be more intuitive to consider. It would therefore be valuable if it could be determined whether the operator aimed to adjust the configuration or just the pose of the tool. For example, a large configuration change yielding a small movement of the tool, should promote the hypothesis that the operator aimed to adjust the configuration.

It should be stated that the scenarios evaluated here are not covering the whole range of plausible scenarios related to this method, and it remains as future work to investigate the generalizability, and user experience, more thoroughly. The last part of the resulting movement is guaranteed to follow the retained part of the corrective demonstration accurately, given enough DMP basis functions. Hence, the only source of error on that part is a faulty demonstration. For instance, the movement might require higher accuracy than what is possible to demonstrate using lead-through programming. Another limitation with this method is that it is difficult for the operator to very accurately determine which part of the faulty trajectory to retain, since this is done autonomously. However, for the experiments performed here, the estimation of the operator was sufficient to demonstrate desired behavior. The benefit with this approach is that it saves time as the operator does not have to specify all details explicitly.

%The work presented here should however serve as a proof of concept, and an idea to build on.

\section{Conclusions}
In this paper, an approach for modification of DMPs, using lead-through programming, was presented. It allowed a robot operator to modify the last part of a faulty generated trajectory, instead of demonstrating a new one from the beginning. Based on the corrective demonstration, modified DMPs were formed automatically. A real-time application, that did not require any additional engineering work by the user, was developed, and verified experimentally. A video showing the functionality is available as a publication attachment, and a version with higher resolution is available on \cite{mod_youtube}.
\label{sec:conclusions}

%\section{Future Work}

%%%%%%%%%%%%%%%%%%%%%%%%%%%%%%%%%%%%%%%%%%%%%%%%%%%%%%%%%%%%%%%%%%%%%%%%%%%%%%%%
%\section*{APPENDIX}

%Appendixes should appear before the acknowledgment.

%\section*{Acknowledgments}

%NOTES FOR BIBLIOGRAPHY:
%! LaTeX Error: Something's wrong--perhaps a missing \item.
%Perhaps changed .bib file
%Remove generated files, include one correct ref (crucial)
%make sure all necessary files still there, like IEEEtran.bst
%include bibtex in build under options -> configure texmaker.
\bibliographystyle{IEEEtran}
\bibliography{icra2017_mod}

\end{document}

%% file: figs/cascade_scheme.tex
\usetikzlibrary{shapes,positioning}
\usetikzlibrary{narrow}

\tikzset{block/.style={draw, rectangle, line width=2pt,
     minimum height=3em, minimum width=3em, outer sep=0pt}}
\tikzset{every picture/.style={auto, line width=1pt,
          >=narrow,font=\small}}
\begin{center}
\begin{tikzpicture}[auto, node distance=1.0cm,>=latex',scale=0.9, every node/.style={scale=0.9}]
\node[text width=2cm,align=center, block,fill=red!20](initdemo){Initial \\ demonstration};
\node[block, right=3.5mm of initdemo](fitdmp){Create DMP};
\node[text width=1.8cm,align=center, block, fill=blue!20, right=10mm of fitdmp](eval){Evaluation};
\coordinate[left=0mm of eval](preeval);
\coordinate[right=6mm of eval](posteval);
\coordinate[right=2mm of posteval](postposteval);
\node[text width=2cm,align=center, block, fill=red!20, below=10mm of posteval](corrdemo){Corrective \\ demonstration};
\node[text width=1cm,align=center, block, below=10mm of preeval](moddmp){Modify \\ DMP};
\node[text width=2cm,align=center, block, fill=blue!20, above=9mm of posteval](reinf){Further \\ improvement};

\node (succ) [above=3.2mm of postposteval]{Successful};
\node (unsucc) [below=4.5mm of postposteval]{Unsuccessful};
\draw[->](initdemo)--(fitdmp);
\draw[->](corrdemo)--(moddmp);
\draw[->](fitdmp)--(eval);
\draw[->](eval)--(reinf);
\draw[->](eval)--(corrdemo);
\draw[->](moddmp)--(eval);

%Draw the large dashed box:
\coordinate[left=2mm of moddmp](tmp1);
\coordinate[below=6mm of tmp1](lowerleft);
\coordinate[above=28mm of lowerleft](upperleft);
\draw[dashed](lowerleft)--(upperleft);
\coordinate[right=42.5mm of upperleft](upperright);
\draw[dashed](upperleft)--(upperright);
\coordinate[below=28mm of upperright](lowerright);
\draw[dashed](upperright)--(lowerright);
\draw[dashed](lowerright)--(lowerleft);
\end{tikzpicture}
\end{center}

%% file: figs/adjust_notes.tex
\begin{tikzpicture}

\begin{axis}[%
width=\figurewidth,
height=\figureheight,
scale only axis,
xmin=100,
xmax=140,
xlabel={x [mm]},
xmajorgrids,
ymin=0,
ymax=50,
ylabel={z [mm]},
ymajorgrids,
legend style={draw=black,fill=white,legend cell align=left}
]
\addplot [color=white!50!red,solid,line width=5.0pt]
  table[row sep=crcr]{%
0	140.463091378\\
0.00308358299997735	140.457144087\\
0.0203699589999928	140.393491524\\
0.140442180999969	140.304630024\\
0.701843286999974	140.594079901\\
1.65746168099997	141.265330598\\
2.745038991	141.818598007\\
3.64899754999999	141.993352987\\
4.49078690299996	141.855011206\\
5.48085534299997	141.494089601\\
6.63339217699996	141.066894663\\
7.84344388899996	140.573978829\\
8.98671143199999	140.119895539\\
10.1215486	139.564753132\\
11.267604605	138.83976658\\
12.377478531	137.644423284\\
13.695679894	136.366455469\\
15.147944397	135.158795031\\
16.658526914	134.213695994\\
18.204950258	133.079319887\\
19.766242667	131.908011543\\
21.165397277	130.755744852\\
22.412003624	129.687565062\\
23.593952599	128.81713784\\
24.723244586	128.114076475\\
25.791991447	127.536933495\\
26.896457521	127.055796447\\
28.054210981	126.351845515\\
29.255270749	125.409557757\\
30.629639499	124.31677634\\
32.238966815	122.994514726\\
34.116160245	121.435917863\\
36.004673314	119.735449318\\
37.904143458	117.953023831\\
39.748686065	115.856098276\\
41.511183463	113.967855385\\
43.640898633	111.859602507\\
45.798510669	109.958115579\\
47.89087894	108.091379648\\
49.901278971	106.335849801\\
51.842690536	104.41272511\\
53.619793135	102.582581146\\
55.319071289	100.84745537\\
57.055117286	99.289226351\\
58.764178571	97.803690998\\
60.407900916	96.397761477\\
62.168482104	94.773690652\\
64.022475953	92.707247072\\
65.725969646	90.39483361\\
67.421715631	87.76151879\\
69.386868281	85.109482587\\
71.265634725	82.369057937\\
73.037336999	79.375572928\\
74.869740899	76.152954747\\
76.751255319	73.391965398\\
78.49096103	71.206121781\\
80.206745746	69.104941463\\
81.70867415	67.278573587\\
82.975999917	65.397408884\\
84.144989987	63.239218078\\
85.339230101	61.156261553\\
86.438117602	59.2818322\\
87.417763935	57.457831428\\
88.336582815	55.602149711\\
89.305559714	53.865414481\\
90.363195545	52.335772\\
91.518328297	50.705084892\\
92.762988002	49.13314552\\
94.011306179	47.463430326\\
95.124689592	46.027291838\\
95.946796111	44.889030087\\
96.891294798	43.997309523\\
97.929754621	43.189779511\\
98.928516113	42.331683426\\
99.856037211	41.339081309\\
100.787802673	39.820476724\\
101.659157692	37.976457853\\
102.393521154	36.327219163\\
103.040569741	34.885437686\\
103.634615612	33.513403313\\
104.284363059	32.009457176\\
105.126321255	30.40686201\\
106.015876307	28.864924809\\
106.695638901	27.428275176\\
107.187585637	26.311787104\\
107.850251501	25.783046599\\
108.851260032	25.458293444\\
109.619564484	25.073508086\\
110.311316443	24.489754482\\
110.881495264	23.6236068\\
111.344825923	22.627049989\\
111.811739109	21.746840389\\
112.180791495	20.384286332\\
112.440223779	18.738008878\\
112.544495364	17.006782822\\
112.675978208	15.468238066\\
112.898936644	14.115265447\\
113.193910132	13.031951655\\
113.523546309	12.014580772\\
113.752554988	11.500898599\\
113.873681429	11.173458306\\
114.000789858	10.847273075\\
114.231613013	10.608987533\\
114.564403006	10.105795715\\
115.17909576	8.69745061500001\\
115.656806042	6.49471937499999\\
116.428858987	6.70908925000001\\
116.73453575	6.554459622\\
116.854745781	6.358458644\\
116.894488202	6.33634570500001\\
116.905629789	6.34236913399999\\
};
\addlegendentry{Deficient trajectory};

\addplot [color=red,solid,line width=5.0pt]
  table[row sep=crcr]{%
117.184569534	6.70581914100001\\
117.186539041	6.70670031500001\\
117.190862834	6.71251994400001\\
117.254504573	6.783694156\\
117.329376858	6.87094492400001\\
117.350802766	6.89471633700001\\
117.388538625	6.93600898100001\\
117.508353619	7.07195814299999\\
117.63955769	7.22424777500001\\
117.749100011	7.34907459199999\\
117.891576447	7.51460970400001\\
118.09558241	7.749950867\\
118.326454673	8.109880175\\
118.542299275	9.02013981600001\\
118.734087203	10.078172848\\
118.870170797	10.901255786\\
118.982822641	11.423839612\\
119.027002512	11.586174246\\
119.041771586	11.639647081\\
119.033814289	11.666255525\\
119.030098316	11.664508087\\
119.024742839	11.679326428\\
119.033917565	11.722601903\\
119.019757005	11.758211689\\
118.923925906	11.727290964\\
118.817707075	11.657115261\\
118.74469525	11.613380039\\
118.71738917	11.65805862\\
118.706237561	11.767228303\\
118.713885088	11.997053914\\
118.791830884	12.23768006\\
118.878727966	12.421693481\\
118.851799626	12.593599458\\
118.808122114	13.084272047\\
118.791376066	14.109558858\\
118.782572304	15.184313235\\
118.676977535	16.05282968\\
118.57320573	16.195012051\\
118.375085407	16.014512153\\
118.183118773	15.895817481\\
118.064160467	15.963747039\\
118.000491676	16.395463188\\
117.964809422	17.010160586\\
117.928771798	17.345132121\\
117.876168033	17.461399945\\
117.81401168	17.540037625\\
117.783125657	17.675166955\\
117.763276705	18.094239467\\
117.700091308	19.066263077\\
117.698589536	19.834755164\\
117.658113202	20.538716046\\
117.596151111	21.116180403\\
117.506068536	21.238350283\\
117.265594472	21.471872317\\
117.139251933	22.312386094\\
117.129605661	23.476059961\\
117.061465285	24.719556387\\
117.127689291	25.642733797\\
117.189679661	26.030068524\\
117.273567083	26.154182276\\
117.395564685	26.316505963\\
117.4626943	26.436048973\\
117.470108799	26.492675382\\
117.474266808	26.554619382\\
117.475102729	26.57857974\\
117.481396679	26.595872598\\
117.485346083	26.607079745\\
117.48845576	26.610855433\\
117.503570799	26.621789684\\
117.52804667	26.637354317\\
117.560351775	26.661575023\\
117.639977552	26.721154456\\
117.759534573	26.806627837\\
117.912648456	26.923650125\\
118.131271356	27.113172811\\
118.486342398	27.336752398\\
118.766829438	27.479731519\\
118.950615331	27.392958336\\
119.076573333	27.262195217\\
119.255665376	26.866287165\\
119.474758165	26.163188143\\
119.67876187	25.876659928\\
119.854092169	25.795255525\\
120.039716923	25.761539616\\
120.395281402	25.73492019\\
120.736156616	25.585302033\\
120.937478034	25.352014469\\
121.140877832	25.08267479\\
121.398095543	24.531345722\\
121.759755181	23.89448289\\
122.280203258	23.277685061\\
122.873839209	22.756737628\\
123.452293817	22.354037049\\
124.034807949	22.067471211\\
124.366169658	21.99322345\\
124.622739233	21.732265378\\
124.92464031	21.19422189\\
125.199312599	21.04919097\\
125.428512783	21.186171855\\
125.554733799	21.04241437\\
125.711556647	20.221554981\\
125.892092202	19.304664708\\
126.12761019	18.315338423\\
126.313078715	17.597090207\\
126.474500633	17.272142886\\
126.684518943	16.894069158\\
126.961012846	16.561435299\\
127.080089497	16.105559329\\
127.103296197	15.528138464\\
127.124556412	15.201367179\\
127.176827048	15.151623275\\
127.248720136	15.103513662\\
127.363549548	15.077526579\\
127.555096549	14.981118139\\
127.953045636	14.842804711\\
128.309796794	14.713757429\\
128.559625166	14.449846979\\
128.692937325	14.091784547\\
128.858030629	13.507768025\\
129.074329394	13.095839801\\
129.251740947	12.955957158\\
129.334074163	12.8559682\\
129.426141082	12.691746368\\
129.548085	12.132796296\\
129.68241251	11.589893579\\
129.856434665	11.460557644\\
130.094557026	11.082701668\\
130.356087757	10.388417439\\
130.480069214	10.269245413\\
130.535525101	10.303407794\\
130.602034832	10.284234625\\
130.705614124	10.13912126\\
130.799539906	9.57184424600001\\
130.857640955	8.94036862199999\\
130.896037498	8.86462779999999\\
130.940801659	8.90056696299999\\
131.12760065	9.09238921900001\\
131.237330279	9.17781382500002\\
131.253739215	9.099987505\\
131.287052087	9.063596719\\
131.331307994	8.99944108700001\\
131.385130277	8.57381211699999\\
131.497384446	7.69759403700002\\
131.544306499	7.00061264300001\\
131.631175365	6.424256273\\
131.758577946	6.202933586\\
132.019287385	6.407178335\\
132.298651087	6.52634217600001\\
132.544912172	6.53916063599999\\
132.751855897	6.53335232399999\\
132.828184206	6.316565704\\
132.959412204	6.05255532000001\\
133.183521235	5.94937842600001\\
133.449883345	5.88890010599999\\
133.696777041	5.781465819\\
133.943837981	5.51419529200001\\
134.087565254	5.17151089399999\\
134.120510756	4.92702400799999\\
134.145156896	4.87967854800002\\
134.155480938	4.834942404\\
134.153586084	4.73406413800001\\
134.134256308	4.33591867999999\\
134.102760049	3.70857953699999\\
134.082332068	2.96736787399999\\
134.083850736	2.482875439\\
134.100530564	2.19807138499999\\
134.119829326	1.88030407100001\\
134.128865424	1.43215140699999\\
134.148486923	1.16901198900001\\
134.175901836	1.06403677\\
134.194145427	1.029691235\\
134.206260002	0.994478119999997\\
134.219185404	0.925387341000004\\
134.219217093	0.618446831999989\\
134.26485851	0.0850757810000005\\
134.366105526	0\\
};
\addlegendentry{Corrective trajectory};
\end{axis}

\draw [|-|, line width=.5mm] (2,2.1) -- (2.8,2.6);
\node at (2.4,2.6) {$d_m$};;

\end{tikzpicture}%

%% file: figs/adjust_zoom.tex
\begin{tikzpicture}

\begin{axis}[%
width=\figurewidth,
height=\figureheight,
scale only axis,
xmin=100,
xmax=140,
xlabel={x [mm]},
xmajorgrids,
ymin=0,
ymax=50,
ylabel={z [mm]},
ymajorgrids,
legend style={draw=black,fill=white,legend cell align=left}
]
\addplot [color=white!50!red,solid,line width=5.0pt]
  table[row sep=crcr]{%
0	140.463091378\\
0.00308358299997735	140.457144087\\
0.0203699589999928	140.393491524\\
0.140442180999969	140.304630024\\
0.701843286999974	140.594079901\\
1.65746168099997	141.265330598\\
2.745038991	141.818598007\\
3.64899754999999	141.993352987\\
4.49078690299996	141.855011206\\
5.48085534299997	141.494089601\\
6.63339217699996	141.066894663\\
7.84344388899996	140.573978829\\
8.98671143199999	140.119895539\\
10.1215486	139.564753132\\
11.267604605	138.83976658\\
12.377478531	137.644423284\\
13.695679894	136.366455469\\
15.147944397	135.158795031\\
16.658526914	134.213695994\\
18.204950258	133.079319887\\
19.766242667	131.908011543\\
21.165397277	130.755744852\\
22.412003624	129.687565062\\
23.593952599	128.81713784\\
24.723244586	128.114076475\\
25.791991447	127.536933495\\
26.896457521	127.055796447\\
28.054210981	126.351845515\\
29.255270749	125.409557757\\
30.629639499	124.31677634\\
32.238966815	122.994514726\\
34.116160245	121.435917863\\
36.004673314	119.735449318\\
37.904143458	117.953023831\\
39.748686065	115.856098276\\
41.511183463	113.967855385\\
43.640898633	111.859602507\\
45.798510669	109.958115579\\
47.89087894	108.091379648\\
49.901278971	106.335849801\\
51.842690536	104.41272511\\
53.619793135	102.582581146\\
55.319071289	100.84745537\\
57.055117286	99.289226351\\
58.764178571	97.803690998\\
60.407900916	96.397761477\\
62.168482104	94.773690652\\
64.022475953	92.707247072\\
65.725969646	90.39483361\\
67.421715631	87.76151879\\
69.386868281	85.109482587\\
71.265634725	82.369057937\\
73.037336999	79.375572928\\
74.869740899	76.152954747\\
76.751255319	73.391965398\\
78.49096103	71.206121781\\
80.206745746	69.104941463\\
81.70867415	67.278573587\\
82.975999917	65.397408884\\
84.144989987	63.239218078\\
85.339230101	61.156261553\\
86.438117602	59.2818322\\
87.417763935	57.457831428\\
88.336582815	55.602149711\\
89.305559714	53.865414481\\
90.363195545	52.335772\\
91.518328297	50.705084892\\
92.762988002	49.13314552\\
94.011306179	47.463430326\\
95.124689592	46.027291838\\
95.946796111	44.889030087\\
96.891294798	43.997309523\\
97.929754621	43.189779511\\
98.928516113	42.331683426\\
99.856037211	41.339081309\\
100.787802673	39.820476724\\
101.659157692	37.976457853\\
102.393521154	36.327219163\\
103.040569741	34.885437686\\
103.634615612	33.513403313\\
104.284363059	32.009457176\\
105.126321255	30.40686201\\
106.015876307	28.864924809\\
106.695638901	27.428275176\\
107.187585637	26.311787104\\
107.850251501	25.783046599\\
108.851260032	25.458293444\\
109.619564484	25.073508086\\
110.311316443	24.489754482\\
110.881495264	23.6236068\\
111.344825923	22.627049989\\
111.811739109	21.746840389\\
112.180791495	20.384286332\\
112.440223779	18.738008878\\
112.544495364	17.006782822\\
112.675978208	15.468238066\\
112.898936644	14.115265447\\
113.193910132	13.031951655\\
113.523546309	12.014580772\\
113.752554988	11.500898599\\
113.873681429	11.173458306\\
114.000789858	10.847273075\\
114.231613013	10.608987533\\
114.564403006	10.105795715\\
115.17909576	8.69745061500001\\
115.656806042	6.49471937499999\\
116.428858987	6.70908925000001\\
116.73453575	6.554459622\\
116.854745781	6.358458644\\
116.894488202	6.33634570500001\\
116.905629789	6.34236913399999\\
};
\addlegendentry{Deficient trajectory};

\addplot [color=red,solid,line width=5.0pt]
  table[row sep=crcr]{%
117.184569534	6.70581914100001\\
117.186539041	6.70670031500001\\
117.190862834	6.71251994400001\\
117.254504573	6.783694156\\
117.329376858	6.87094492400001\\
117.350802766	6.89471633700001\\
117.388538625	6.93600898100001\\
117.508353619	7.07195814299999\\
117.63955769	7.22424777500001\\
117.749100011	7.34907459199999\\
117.891576447	7.51460970400001\\
118.09558241	7.749950867\\
118.326454673	8.109880175\\
118.542299275	9.02013981600001\\
118.734087203	10.078172848\\
118.870170797	10.901255786\\
118.982822641	11.423839612\\
119.027002512	11.586174246\\
119.041771586	11.639647081\\
119.033814289	11.666255525\\
119.030098316	11.664508087\\
119.024742839	11.679326428\\
119.033917565	11.722601903\\
119.019757005	11.758211689\\
118.923925906	11.727290964\\
118.817707075	11.657115261\\
118.74469525	11.613380039\\
118.71738917	11.65805862\\
118.706237561	11.767228303\\
118.713885088	11.997053914\\
118.791830884	12.23768006\\
118.878727966	12.421693481\\
118.851799626	12.593599458\\
118.808122114	13.084272047\\
118.791376066	14.109558858\\
118.782572304	15.184313235\\
118.676977535	16.05282968\\
118.57320573	16.195012051\\
118.375085407	16.014512153\\
118.183118773	15.895817481\\
118.064160467	15.963747039\\
118.000491676	16.395463188\\
117.964809422	17.010160586\\
117.928771798	17.345132121\\
117.876168033	17.461399945\\
117.81401168	17.540037625\\
117.783125657	17.675166955\\
117.763276705	18.094239467\\
117.700091308	19.066263077\\
117.698589536	19.834755164\\
117.658113202	20.538716046\\
117.596151111	21.116180403\\
117.506068536	21.238350283\\
117.265594472	21.471872317\\
117.139251933	22.312386094\\
117.129605661	23.476059961\\
117.061465285	24.719556387\\
117.127689291	25.642733797\\
117.189679661	26.030068524\\
117.273567083	26.154182276\\
117.395564685	26.316505963\\
117.4626943	26.436048973\\
117.470108799	26.492675382\\
117.474266808	26.554619382\\
117.475102729	26.57857974\\
117.481396679	26.595872598\\
117.485346083	26.607079745\\
117.48845576	26.610855433\\
117.503570799	26.621789684\\
117.52804667	26.637354317\\
117.560351775	26.661575023\\
117.639977552	26.721154456\\
117.759534573	26.806627837\\
117.912648456	26.923650125\\
118.131271356	27.113172811\\
118.486342398	27.336752398\\
118.766829438	27.479731519\\
118.950615331	27.392958336\\
119.076573333	27.262195217\\
119.255665376	26.866287165\\
119.474758165	26.163188143\\
119.67876187	25.876659928\\
119.854092169	25.795255525\\
120.039716923	25.761539616\\
120.395281402	25.73492019\\
120.736156616	25.585302033\\
120.937478034	25.352014469\\
121.140877832	25.08267479\\
121.398095543	24.531345722\\
121.759755181	23.89448289\\
122.280203258	23.277685061\\
122.873839209	22.756737628\\
123.452293817	22.354037049\\
124.034807949	22.067471211\\
124.366169658	21.99322345\\
124.622739233	21.732265378\\
124.92464031	21.19422189\\
125.199312599	21.04919097\\
125.428512783	21.186171855\\
125.554733799	21.04241437\\
125.711556647	20.221554981\\
125.892092202	19.304664708\\
126.12761019	18.315338423\\
126.313078715	17.597090207\\
126.474500633	17.272142886\\
126.684518943	16.894069158\\
126.961012846	16.561435299\\
127.080089497	16.105559329\\
127.103296197	15.528138464\\
127.124556412	15.201367179\\
127.176827048	15.151623275\\
127.248720136	15.103513662\\
127.363549548	15.077526579\\
127.555096549	14.981118139\\
127.953045636	14.842804711\\
128.309796794	14.713757429\\
128.559625166	14.449846979\\
128.692937325	14.091784547\\
128.858030629	13.507768025\\
129.074329394	13.095839801\\
129.251740947	12.955957158\\
129.334074163	12.8559682\\
129.426141082	12.691746368\\
129.548085	12.132796296\\
129.68241251	11.589893579\\
129.856434665	11.460557644\\
130.094557026	11.082701668\\
130.356087757	10.388417439\\
130.480069214	10.269245413\\
130.535525101	10.303407794\\
130.602034832	10.284234625\\
130.705614124	10.13912126\\
130.799539906	9.57184424600001\\
130.857640955	8.94036862199999\\
130.896037498	8.86462779999999\\
130.940801659	8.90056696299999\\
131.12760065	9.09238921900001\\
131.237330279	9.17781382500002\\
131.253739215	9.099987505\\
131.287052087	9.063596719\\
131.331307994	8.99944108700001\\
131.385130277	8.57381211699999\\
131.497384446	7.69759403700002\\
131.544306499	7.00061264300001\\
131.631175365	6.424256273\\
131.758577946	6.202933586\\
132.019287385	6.407178335\\
132.298651087	6.52634217600001\\
132.544912172	6.53916063599999\\
132.751855897	6.53335232399999\\
132.828184206	6.316565704\\
132.959412204	6.05255532000001\\
133.183521235	5.94937842600001\\
133.449883345	5.88890010599999\\
133.696777041	5.781465819\\
133.943837981	5.51419529200001\\
134.087565254	5.17151089399999\\
134.120510756	4.92702400799999\\
134.145156896	4.87967854800002\\
134.155480938	4.834942404\\
134.153586084	4.73406413800001\\
134.134256308	4.33591867999999\\
134.102760049	3.70857953699999\\
134.082332068	2.96736787399999\\
134.083850736	2.482875439\\
134.100530564	2.19807138499999\\
134.119829326	1.88030407100001\\
134.128865424	1.43215140699999\\
134.148486923	1.16901198900001\\
134.175901836	1.06403677\\
134.194145427	1.029691235\\
134.206260002	0.994478119999997\\
134.219185404	0.925387341000004\\
134.219217093	0.618446831999989\\
134.26485851	0.0850757810000005\\
134.366105526	0\\
};
\addlegendentry{Corrective trajectory};

\addplot [color=black,solid,line width=2.0pt]
  table[row sep=crcr]{%
0	140.463091378\\
0.00308358299997735	140.457144087\\
0.0203699589999928	140.393491524\\
0.140442180999969	140.304630024\\
0.701843286999974	140.594079901\\
1.65746168099997	141.265330598\\
2.745038991	141.818598007\\
3.64899754999999	141.993352987\\
4.49078690299996	141.855011206\\
5.48085534299997	141.494089601\\
6.63339217699996	141.066894663\\
7.84344388899996	140.573978829\\
8.98671143199999	140.119895539\\
10.1215486	139.564753132\\
11.267604605	138.83976658\\
12.377478531	137.644423284\\
13.695679894	136.366455469\\
15.147944397	135.158795031\\
16.658526914	134.213695994\\
18.204950258	133.079319887\\
19.766242667	131.908011543\\
21.165397277	130.755744852\\
22.412003624	129.687565062\\
23.593952599	128.81713784\\
24.723244586	128.114076475\\
25.791991447	127.536933495\\
26.896457521	127.055796447\\
28.054210981	126.351845515\\
29.255270749	125.409557757\\
30.629639499	124.31677634\\
32.238966815	122.994514726\\
34.116160245	121.435917863\\
36.004673314	119.735449318\\
37.904143458	117.953023831\\
39.748686065	115.856098276\\
41.511183463	113.967855385\\
43.640898633	111.859602507\\
45.798510669	109.958115579\\
47.89087894	108.091379648\\
49.901278971	106.335849801\\
51.842690536	104.41272511\\
53.619793135	102.582581146\\
55.319071289	100.84745537\\
57.055117286	99.289226351\\
58.764178571	97.803690998\\
60.407900916	96.397761477\\
62.168482104	94.773690652\\
64.022475953	92.707247072\\
65.725969646	90.39483361\\
67.421715631	87.76151879\\
69.386868281	85.109482587\\
71.265634725	82.369057937\\
73.037336999	79.375572928\\
74.869740899	76.152954747\\
76.751255319	73.391965398\\
78.49096103	71.206121781\\
80.206745746	69.104941463\\
81.70867415	67.278573587\\
82.975999917	65.397408884\\
84.144989987	63.239218078\\
85.339230101	61.156261553\\
86.438117602	59.2818322\\
87.417763935	57.457831428\\
88.336582815	55.602149711\\
89.305559714	53.865414481\\
90.363195545	52.335772\\
91.518328297	50.705084892\\
92.762988002	49.13314552\\
94.011306179	47.463430326\\
95.124689592	46.027291838\\
95.946796111	44.889030087\\
96.891294798	43.997309523\\
97.929754621	43.189779511\\
98.928516113	42.331683426\\
99.856037211	41.339081309\\
100.620633154009	39.3801307974991\\
101.370116849094	37.8022755854935\\
102.074507285708	36.1628327517566\\
102.774663071479	34.5205856100479\\
103.520065003802	32.9223836780777\\
104.356748998797	31.4120906795625\\
105.316929830884	30.0346241498069\\
106.414570671058	28.8304713122816\\
107.648080247925	27.8279206427407\\
109.001697613279	27.0388532064176\\
110.441755221246	26.4577359253435\\
111.913740915068	26.0651473760213\\
113.34682900972	25.8353122323324\\
114.663497139028	25.7409071356765\\
115.785087010621	25.7545031388137\\
116.636768567055	25.8480683596705\\
117.151569480973	25.991858588918\\
117.27356708297	26.1541822758714\\
};
\addlegendentry{Modified trajectory};

\addplot [color=blue,line width=3.0pt,mark size=6.0pt,only marks,mark=x,mark options={solid}]
  table[row sep=crcr]{%
117.273567083	26.154182276\\
};
\addlegendentry{Separation point set by user};

\addplot [color=green,line width=3.0pt,mark size=6.0pt,only marks,mark=x,mark options={solid}]
  table[row sep=crcr]{%
112.180791495	20.384286332\\
};
\addlegendentry{Separation point set automatically};

\addplot [color=black,solid,line width=2.0pt,forget plot]
  table[row sep=crcr]{%
117.273567083	26.154182276\\
117.395564685	26.316505963\\
117.4626943	26.436048973\\
117.470108799	26.492675382\\
117.474266808	26.554619382\\
117.475102729	26.57857974\\
117.481396679	26.595872598\\
117.485346083	26.607079745\\
117.48845576	26.610855433\\
117.503570799	26.621789684\\
117.52804667	26.637354317\\
117.560351775	26.661575023\\
117.639977552	26.721154456\\
117.759534573	26.806627837\\
117.912648456	26.923650125\\
118.131271356	27.113172811\\
118.486342398	27.336752398\\
118.766829438	27.479731519\\
118.950615331	27.392958336\\
119.076573333	27.262195217\\
119.255665376	26.866287165\\
119.474758165	26.163188143\\
119.67876187	25.876659928\\
119.854092169	25.795255525\\
120.039716923	25.761539616\\
120.395281402	25.73492019\\
120.736156616	25.585302033\\
120.937478034	25.352014469\\
121.140877832	25.08267479\\
121.398095543	24.531345722\\
121.759755181	23.89448289\\
122.280203258	23.277685061\\
122.873839209	22.756737628\\
123.452293817	22.354037049\\
124.034807949	22.067471211\\
124.366169658	21.99322345\\
124.622739233	21.732265378\\
124.92464031	21.19422189\\
125.199312599	21.04919097\\
125.428512783	21.186171855\\
125.554733799	21.04241437\\
125.711556647	20.221554981\\
125.892092202	19.304664708\\
126.12761019	18.315338423\\
126.313078715	17.597090207\\
126.474500633	17.272142886\\
126.684518943	16.894069158\\
126.961012846	16.561435299\\
127.080089497	16.105559329\\
127.103296197	15.528138464\\
127.124556412	15.201367179\\
127.176827048	15.151623275\\
127.248720136	15.103513662\\
127.363549548	15.077526579\\
127.555096549	14.981118139\\
127.953045636	14.842804711\\
128.309796794	14.713757429\\
128.559625166	14.449846979\\
128.692937325	14.091784547\\
128.858030629	13.507768025\\
129.074329394	13.095839801\\
129.251740947	12.955957158\\
129.334074163	12.8559682\\
129.426141082	12.691746368\\
129.548085	12.132796296\\
129.68241251	11.589893579\\
129.856434665	11.460557644\\
130.094557026	11.082701668\\
130.356087757	10.388417439\\
130.480069214	10.269245413\\
130.535525101	10.303407794\\
130.602034832	10.284234625\\
130.705614124	10.13912126\\
130.799539906	9.57184424600001\\
130.857640955	8.94036862199999\\
130.896037498	8.86462779999999\\
130.940801659	8.90056696299999\\
131.12760065	9.09238921900001\\
131.237330279	9.17781382500002\\
131.253739215	9.099987505\\
131.287052087	9.063596719\\
131.331307994	8.99944108700001\\
131.385130277	8.57381211699999\\
131.497384446	7.69759403700002\\
131.544306499	7.00061264300001\\
131.631175365	6.424256273\\
131.758577946	6.202933586\\
132.019287385	6.407178335\\
132.298651087	6.52634217600001\\
132.544912172	6.53916063599999\\
132.751855897	6.53335232399999\\
132.828184206	6.316565704\\
132.959412204	6.05255532000001\\
133.183521235	5.94937842600001\\
133.449883345	5.88890010599999\\
133.696777041	5.781465819\\
133.943837981	5.51419529200001\\
134.087565254	5.17151089399999\\
134.120510756	4.92702400799999\\
134.145156896	4.87967854800002\\
134.155480938	4.834942404\\
134.153586084	4.73406413800001\\
134.134256308	4.33591867999999\\
134.102760049	3.70857953699999\\
134.082332068	2.96736787399999\\
134.083850736	2.482875439\\
134.100530564	2.19807138499999\\
134.119829326	1.88030407100001\\
134.128865424	1.43215140699999\\
134.148486923	1.16901198900001\\
134.175901836	1.06403677\\
134.194145427	1.029691235\\
134.206260002	0.994478119999997\\
134.219185404	0.925387341000004\\
134.219217093	0.618446831999989\\
134.26485851	0.0850757810000005\\
134.366105526	0\\
};
\end{axis}

\draw [->] (.35,4) -- (1,3.2);
\draw [->] (3.3,1) -- (3.1,2);
\draw [->] (4.4,2) -- (5.4,1);

\end{tikzpicture}%

%% file: figs/place_a1.tex
% This file was created by matlab2tikz v0.4.7 running on MATLAB 7.14.
% Copyright (c) 2008--2014, Nico Schlömer <nico.schloemer@gmail.com>
% All rights reserved.
% Minimal pgfplots version: 1.3
% 
% The latest updates can be retrieved from
%   http://www.mathworks.com/matlabcentral/fileexchange/22022-matlab2tikz
% where you can also make suggestions and rate matlab2tikz.
% 
\begin{tikzpicture}

\begin{axis}[%
width=\figurewidth,
height=\figureheight,
scale only axis,
xmin=10.9775375467,
xmax=275,
xmajorgrids,
ymajorgrids,
xlabel={y [mm]},
ymin=28.4880469855083,
ymax=222.168397523992,
ylabel={z [mm]},
legend pos=north west,
legend style={draw=black,fill=white,legend cell align=left}
]
\addplot [color=white!50!red,solid,line width=5.0pt]
  table[row sep=crcr]{%
256.543626164	213.6734180074\\
243.327411128	210.8414816094\\
217.61217637	204.7677834682\\
188.585181383	196.557414432\\
161.411168318	187.7637455539\\
137.547225384	178.9417460599\\
116.795559654	170.1644688401\\
97.1102754626	159.6792977718\\
77.806137912	147.007751438\\
59.4670445872	130.881550626\\
44.6395492476	113.823093874\\
33.5855511479	99.498715294\\
25.2583739714	88.184954899\\
18.5070571046	78.794129778\\
14.4003228378	70.885169143\\
12.7864500787	64.824791369\\
12.1203985858	60.602786921\\
11.6870184766	58.625051615\\
11.3775710178	58.158323288\\
};
\addlegendentry{Deficient trajectory};

\addplot [color=red,solid,line width=5.0pt]
  table[row sep=crcr]{%
11.2162126985	60.542704133\\
10.9775375467	61.004776456\\
11.6413370186	63.344204566\\
14.08000133	64.780517759\\
19.2580929959	69.72450528\\
22.1714408975	74.660897932\\
25.8863707053	77.892670433\\
27.1558359822	80.036206967\\
29.4469109131	81.869594755\\
30.0646075826	82.752191316\\
30.0872806027	82.842625264\\
30.1063987672	82.78247364\\
30.1531947889	82.585955904\\
29.8883135453	81.355234426\\
29.0756176715	79.350369492\\
29.1965378178	77.333093204\\
28.1606346533	75.012654168\\
28.0640585415	71.659288929\\
27.4419733446	68.851053851\\
26.7708789603	66.618206299\\
25.9340576408	63.124079665\\
24.302701389	59.829593642\\
23.7103348714	57.852392804\\
23.4207051218	53.830939179\\
23.9014041346	48.863948706\\
24.0446392219	42.493501433\\
23.7977728238	37.867857274\\
23.4674894647	36.817308077\\
};
\addlegendentry{Corrective trajectory};

\addplot [color=blue,solid,line width=2.0pt]
  table[row sep=crcr]{%
255.90866926	213.8391364325\\
245.490502729	210.9316240674\\
225.957957911	205.4027569448\\
200.198815945	198.3167203094\\
173.782740354	190.7704545354\\
150.860530792	183.5784736263\\
131.957419355	176.6249019454\\
115.437648226	169.4613129957\\
99.804847042	161.0738669212\\
83.6761778675	150.9234517681\\
67.3857008714	138.244593438\\
52.2981249177	123.82766454\\
40.23961295	109.339937408\\
31.8130663385	95.552775392\\
26.9043072166	81.799131695\\
24.8286442938	70.687003206\\
25.0291033037	65.565806386\\
25.5947720171	65.328032179\\
25.7157303918	63.600630381\\
24.9565108143	61.10169478\\
24.0065905651	58.643777007\\
23.2672096267	56.180846696\\
23.0462084745	53.276865538\\
23.4902102015	49.110300772\\
23.9517280964	43.97395915\\
23.6228083195	39.410597761\\
23.8515416632	37.34942553\\
};
\addlegendentry{Resulting trajectory};

\draw [->] (235,201) -- (170,181);

\end{axis}
\end{tikzpicture}%

%% file: figs/place_a2.tex
% This file was created by matlab2tikz v0.4.7 running on MATLAB 7.14.
% Copyright (c) 2008--2014, Nico Schlömer <nico.schloemer@gmail.com>
% All rights reserved.
% Minimal pgfplots version: 1.3
% 
% The latest updates can be retrieved from
%   http://www.mathworks.com/matlabcentral/fileexchange/22022-matlab2tikz
% where you can also make suggestions and rate matlab2tikz.
% 
\begin{tikzpicture}

\begin{axis}[%
width=\figurewidth,
height=\figureheight,
scale only axis,
xmin=18.2737362278,
xmax=275,
xmajorgrids,
ymajorgrids,
xlabel={y [mm]},
ymin=27.7541896190986,
ymax=219.178118418001,
ylabel={z [mm]},
legend pos=north west,
legend style={draw=black,fill=white,legend cell align=left}
]
\addplot [color=white!50!red,solid,line width=5.0pt]
  table[row sep=crcr]{%
260.775515939	209.0359950458\\
255.311155056	207.8215122446\\
235.182992208	204.2495227445\\
204.481755128	198.1747951035\\
172.476973689	189.9708361185\\
144.728980149	180.2941092231\\
122.549469464	169.8449620442\\
103.424983016	158.8131587019\\
85.7091749471	145.947368014\\
70.9185419217	132.331290603\\
59.8990163847	118.24360018\\
52.3923349609	103.666716218\\
46.7448074255	87.919585178\\
43.459486237	73.776406001\\
41.9069169765	64.035905701\\
\\
};
\addlegendentry{Deficient trajectory};

\addplot [color=red,solid,line width=5.0pt]
  table[row sep=crcr]{%
40.9870088108	61.659336576\\
40.987532389	61.657114616\\
40.990954108	61.659463946\\
40.9983687297	61.667895477\\
41.0117794844	61.673190687\\
41.4088087302	64.318594581\\
44.1324861488	67.706976583\\
44.7550713699	71.398899094\\
47.1887814392	75.272118954\\
51.0340687182	79.155465072\\
54.5352410525	82.504724865\\
57.6089959409	86.498569948\\
59.6890097323	88.417646428\\
59.6023779592	88.603062455\\
59.6157143467	88.594868485\\
59.6189433401	88.586454449\\
59.628261647	88.568439509\\
59.5992410428	88.562649926\\
59.4464757457	88.581728554\\
57.9538964321	89.947150703\\
53.6466199013	90.674228024\\
48.3288670356	91.364532919\\
41.5458970382	91.016124909\\
36.4261736648	87.826396749\\
32.9848159597	84.516038448\\
30.3967501537	78.318012908\\
27.5221143891	72.578096484\\
25.9168463456	67.189832658\\
25.1703577238	61.173488326\\
24.0575596972	57.909625488\\
21.9210099949	55.478314344\\
19.8754822623	51.695623307\\
18.2737362278	47.056463165\\
18.4318495016	44.18328022\\
20.3229438426	37.894213873\\
};
\addlegendentry{Corrective trajectory};

\addplot [color=blue,solid,line width=2.0pt]
  table[row sep=crcr]{%
260.97717017	208.9734724567\\
259.327483559	208.7054163008\\
247.546100225	206.4009586726\\
226.299573333	201.8941980703\\
198.181730669	195.3578930912\\
169.88743422	187.6095651025\\
145.835064451	179.7175449551\\
127.092174452	172.1275581228\\
112.041811383	164.593930684\\
98.766947855	156.1912285397\\
85.7202315329	146.506485135\\
73.0309044427	134.992657338\\
61.6223203737	122.264226846\\
50.6541631934	108.239868679\\
39.1220734789	93.747721993\\
31.096028466	83.702247331\\
28.6216199928	79.439209228\\
28.1566416651	76.671429413\\
26.9228537361	72.054980024\\
25.628199056	67.146282991\\
24.7998829861	62.377539389\\
23.9932208766	58.454277515\\
22.3590902525	55.622940648\\
20.393516298	52.780894881\\
18.7795494648	49.16840099\\
20.3967650102	38.244666645\\
};
\addlegendentry{Resulting trajectory};

\draw [->] (248,198) -- (180,185);
\draw [->] (48,60) -- (68,88);

\end{axis}
\end{tikzpicture}%

%% file: figs/avoid_a1.tex
% This file was created by matlab2tikz v0.4.7 running on MATLAB 7.14.
% Copyright (c) 2008--2014, Nico Schlömer <nico.schloemer@gmail.com>
% All rights reserved.
% Minimal pgfplots version: 1.3
% 
% The latest updates can be retrieved from
%   http://www.mathworks.com/matlabcentral/fileexchange/22022-matlab2tikz
% where you can also make suggestions and rate matlab2tikz.
% 
\begin{tikzpicture}

\begin{axis}[%
width=\figurewidth,
height=\figureheight,
scale only axis,
xmajorgrids,
ymajorgrids,
xmin=-200,
xmax=400,
xlabel={y [mm]},
ymin=30,
ymax=120,
ylabel={z [mm]},
legend style={draw=black,fill=white,legend cell align=left}
]
\addplot [color=white!50!red,solid,line width=5.0pt]
  table[row sep=crcr]{%
336.748093465	74.704676647\\
325.02006008	73.214649502\\
300.883131839	70.007670502\\
269.365063497	67.315893467\\
239.190506578	66.402443469\\
213.649593198	66.657092795\\
193.037274378	68.654630447\\
174.916869911	72.028777842\\
155.452977601	77.635484666\\
133.026860126	86.07461272\\
108.842722669	95.968982884\\
85.0726694917	104.709080803\\
61.9705050259	110.225831188\\
40.2283973731	112.458222942\\
20.8663347026	113.643254317\\
3.76763634336	114.15962358\\
-11.2498315161	113.210159298\\
-24.0953236589	109.060221282\\
-36.4019396114	102.651918485\\
-49.2431309094	95.857556679\\
-62.3923784357	89.579939493\\
-75.8497775393	84.206076732\\
-90.1951205734	79.95954583\\
-105.692670493	76.83034727\\
-120.960774104	74.16122095\\
-134.919712568	72.443522231\\
-147.223100596	71.754638809\\
-157.175655046	71.887677128\\
-163.479803629	72.234084136\\
-165.794650693	72.149324654\\
-165.491595165	71.854641044\\
};
\addlegendentry{Deficient trajectory};

\addplot [color=red,solid,line width=3.0pt]
  table[row sep=crcr]{%
-164.168093636	73.568733941\\
-162.669611459	76.776960542\\
-155.011719657	74.860147259\\
-141.620813875	75.354847243\\
-123.185635928	76.58000853\\
-101.626223589	80.341011441\\
-80.4418546397	85.958161378\\
-59.7490505502	91.612313923\\
-40.6216483059	99.506210842\\
-27.10515921	106.135412502\\
-21.7266127839	107.156615996\\
-19.6422280183	109.174221981\\
-17.3747814785	108.668485411\\
-14.1184042326	106.906764051\\
-6.16986636364	103.258536285\\
5.7819303281	102.617140247\\
15.2498574667	105.122009014\\
23.5602855309	108.918061611\\
30.7819634977	109.393764881\\
32.652644872	108.693617272\\
32.6863156414	108.641202863\\
32.5061025569	108.848854169\\
32.4441064332	108.830516883\\
32.0692019972	108.716917112\\
31.6837388397	108.779628012\\
31.8136639272	108.545927742\\
30.9790621871	109.905098777\\
27.998016175	110.699276623\\
22.9295287149	111.0843779\\
12.7653206589	113.357879568\\
2.40652703378	115.123984381\\
-8.9468248238	117.491474269\\
-21.0247630033	117.196568816\\
-33.6655539225	114.60033323\\
-45.7416673392	110.514417203\\
-57.7049742849	106.867567675\\
-71.4072085756	101.720160084\\
-86.9588550822	98.197646621\\
-104.841617849	95.36872941\\
-123.115681476	90.198326114\\
-139.250918348	84.286014396\\
-148.173994237	78.379148149\\
-154.205332991	71.475499447\\
-156.796854443	64.614679781\\
-159.845563056	60.254374963\\
-161.904945196	55.082173298\\
-162.728902445	49.617382005\\
-163.396723108	44.924524455\\
-163.875058628	38.737457053\\
-164.194377785	35.272364299\\
};
\addlegendentry{Corrective trajectory};

\addplot [color=blue,solid,line width=1pt]
  table[row sep=crcr]{%
335.93946874	74.803364752\\
327.687443693	73.937861775\\
311.012266412	71.889063508\\
286.169844064	68.94197483\\
258.651804325	67.15617382\\
233.363223364	66.60050278\\
213.202698718	66.969741661\\
197.486432438	68.351027367\\
182.82598478	70.630229146\\
166.881073495	74.253433321\\
148.800912578	79.711553641\\
127.881019996	87.291693649\\
105.626302709	95.610754438\\
83.2988548553	103.413611384\\
62.8941430572	109.04752301\\
45.7644074673	111.995577295\\
30.3673378544	113.109199703\\
14.8113267633	113.988134801\\
2.20346406158	115.621310862\\
-5.89413906923	117.071592308\\
-12.0260115364	117.379678248\\
-20.8233416442	117.222427186\\
-31.7397020322	115.5434231\\
-43.2170228671	112.284517056\\
-54.8747243669	108.337527297\\
-67.0511838133	104.17892041\\
-80.0842819309	100.321421909\\
-94.7887738928	96.800789473\\
-110.571010068	93.224190557\\
-126.117531941	88.997300882\\
-139.390793942	84.10018511\\
-148.909421941	78.830433029\\
-154.703961272	72.988624983\\
-158.240294708	67.083345997\\
-160.465888231	61.591180148\\
-162.145074735	56.624833138\\
-163.283862761	51.716487225\\
-163.803264723	46.751251508\\
-164.131488703	41.858002009\\
-164.271837808	37.380804272\\
-164.510999242	34.844254482\\
};
\addlegendentry{Resulting trajectory};

\draw [->] (150,72) -- (100,90);
\draw [->] (-150,58) -- (-150,42);

\end{axis}
\end{tikzpicture}%